# Transferable Expertise for Autonomous Agents via Real-World Case-Based Learning

Zhenyu Ma[1#], Yuyang Song[1#], Chunyi Yang[2#], Jingyi Zhu[2#], Letian Yang[1], Xukai Jiang[1*]

[1]National Glycoengineering Research Center, Shandong University, Qingdao 266237, China. [2]State Key Laboratory of Microbial Technology, Shandong University, Qingdao 266237, China.

[#]*These authors contributed equally to this work*

**\*Correspondence**

Xukai Jiang, Email: xukai.jiang@sdu.edu.cn

**Abstract:**

LLM-based autonomous agents perform well on general reasoning tasks but still struggle to reliably use task structure, key constraints, and prior experience in complex real-world settings. We propose a case-based learning framework that converts experience from past tasks into reusable knowledge assets, allowing agents to transfer prior case experience to new tasks and perform more structured analysis. Unlike methods based mainly on pretrained knowledge or static prompts, our framework emphasizes extracting and reusing task-relevant knowledge, analytical prompts, and operational skills from real cases. We evaluate the method on a unified benchmark of six complex task categories and compare it with Zero-Shot, Few-Shot, Checklist Prompt, and Rule Memory baselines. Results show that our method achieves consistently strong performance across all tasks and matches or outperforms the best baseline in every case, with especially clear gains on more complex tasks. Further analysis shows that the advantage of case-based learning increases with task complexity, and that practical knowledge acquired by one agent can be reused by others. These findings suggest that case-based learning offers a promising path for building professional agents for real-world work.

# 1. Introduction

In recent years, LLM-based autonomous agents have shown strong capabilities in open-ended tasks such as planning, reasoning, and tool use, raising expectations that they may eventually support complex professional work in scientific research, enterprise platform management, biomedical analysis, and software engineering[1,2]. However, despite their general reasoning ability, current agents often fail in real professional settings, especially on tasks that require precise procedural knowledge, accumulated domain experience, and robust error recovery. A key reason is that professional competence cannot be acquired from static text alone. In real research and engineering practice, expertise is built through repeatedly solving real cases, learning from failures, refining strategies, and accumulating transferable experience. Human experts do not become experts simply by reading documentation; they grow through practice. In contrast, most current LLM agent frameworks still treat each task as an isolated event, discarding useful workflows, repair strategies, failure patterns, and practical insights after execution rather than turning them into reusable knowledge[3].

This limitation is especially evident in complex domains such as AI platform management, scientific analysis, and automated system governance, where success often depends on tacit operational knowledge that is difficult to obtain directly from pretraining corpora. Existing methods, including prompting, retrieval-augmented generation, few-shot learning, and memory-based reflection, only partially address this gap[4-6]. To tackle this problem, we propose Case-Based Learning (CBL), a framework in which agents learn from real-world task cases by actively summarizing what they have learned and storing that experience as structured memory. These memories can also be transferred across agents, making experience shareable and inheritable. Specifically, CBL organizes acquired experience into four complementary modules: fixed domain knowledge, system-level prompt constraints, skill templates, and a curriculum scheduler. Together, these modules allow agents to reuse historical experience across tasks and gradually develop more expert-like behavior.

In summary, this work focuses not only on enabling an agent to complete a single task, but on enabling it to learn through practice, grow through cases, and continuously improve through experience transfer.

## 2. Related Work

### 2.1 LLM-based Autonomous Agents

With recent advances in large language models (LLMs) for language understanding and generation, researchers have increasingly explored their potential as agents for solving complex tasks[3,7]. Unlike traditional dialogue systems, autonomous agents must not only generate text, but also plan tasks, use tools, interact with environments, and adapt their behavior based on feedback[1,8].

Frameworks such as ReAct first combined reasoning and acting in an iterative loop, improving performance on complex tasks[9]. Toolformer further showed that models can learn when to call external tools such as calculators, search engines, and APIs[10]. Subsequent work further expanded this direction toward tool-connected and system-level agents capable of orchestrating APIs, models, and external services in more open-ended settings[11-13]. Later systems, including AutoGPT extended this line of work toward multi-step autonomous planning, enabling agents to decompose goals, manage subtasks, and execute long-horizon workflows[14,15]. Benchmarks such as AgentBench and GAIA were then introduced to evaluate agent capabilities in reasoning, tool use, multi-turn interaction, and task completion[16,17].

Although these studies have substantially improved the general task-solving ability of autonomous agents, they remain focused mainly on planning, reasoning, tool use, and interaction within individual tasks[18]. In contrast, much less attention has been paid to a more fundamental question: how agents can accumulate professional experience through long-term practice and gradually develop into domain experts.

## 2.2 Memory Architectures for Agents

To overcome the limitation that agents effectively start from scratch on each new task, a growing body of research has introduced long-term memory mechanisms into LLM-based agents, enabling them to retain past experience and reuse it in future tasks[3,19-21]. Generative Agents proposed the concept of a memory stream, in which past events are stored according to importance, relevance, and recency, and relevant experiences are retrieved for later decision-making. This work showed that persistent memory can substantially improve behavioral consistency and long-term coherence. MemoryBank, inspired by the Ebbinghaus forgetting curve from cognitive psychology, models memory strength with time-based decay, allowing agents to preserve high-value information while gradually discarding less useful content[20]. Reflexion further proposed that agents perform self-reflection after task failure, converting linguistic summaries into feedback for subsequent decisions and thereby improving through failure-driven learning[6]. ExpeL, in turn, attempts to abstract higher-level lessons from past tasks to guide future few-shot reasoning[22].

Although these studies have begun to address experience accumulation, they still have three main limitations[3,22,23]. First, memory content is often not structurally organized: many approaches treat experience as undifferentiated text fragments rather than distinguishing among stable domain knowledge, behavioral constraints, successful skill templates, and learning-order organization. Second, they largely lack a perspective on knowledge transfer: most systems focus on improving a single agent, with limited exploration of whether experience acquired by one agent can be transferred to another. Third, they are rarely driven by real professional cases: most existing memory studies rely on benchmark-task feedback rather than complex real-world work scenarios. To address these gaps, the CBL framework proposed in this paper advances the study of memory in three ways: it decomposes experience into four structured modules, supports cross-agent knowledge transfer, and emphasizes professional growth through real-world cases.

**2.3 Case-Based Reasoning and Experience Reuse**

Case-Based Reasoning (CBR) is a classic research direction in artificial intelligence[24,25]. Its core idea is that, when facing a new problem, a system can retrieve similar past cases and use their solutions to guide current decision-making. Traditional CBR typically involves four steps: retrieving similar cases, reusing existing solutions, revising or adapting them, and retaining the new experience. This idea is also widespread in human expert decision-making. For example, physicians refer to prior clinical cases, engineers draw on historical fault-handling strategies, and researchers summarize lessons from previous experiments.

In recent years, with the development of LLMs, researchers have renewed interest in the value of case-driven learning for agents[1,3,22]. However, most existing work still remains at the level of few-shot examples, demonstration prompting, or retrieval examples[4,5,26]. In other words, historical cases are mainly used as contextual examples, rather than as opportunities for agents to summarize what they have learned through completing those cases.

The essential difference between the Case-Based Learning (CBL) proposed in this paper and traditional few-shot prompting is that our goal is not to present past answers, but to distill the professional knowledge learned from past cases. Put differently, cases are not demonstration texts; they are sources of learning, and experience is transformed into reusable knowledge assets. This makes CBL closer to a mechanism for professional growth rather than simply a form of example retrieval.

**2.4 Context-Augmented Reasoning**

To compensate for the limits of parametric knowledge in LLMs, many studies have adopted context augmentation strategies that dynamically inject external information during inference[27-29]. The most representative approach is Retrieval-Augmented Generation (RAG), which supplements the model's context by retrieving relevant knowledge from external document collections[30]. In addition, few-shot prompting helps models learn task formats through examples, while Chain-of-Thought and its later variants improve the structure and execution of intermediate reasoningf[31-34].

Our work differs from RAG-style approaches in one key respect: RAG retrieves external documents, whereas CBL reuses practical experience acquired by the agent itself[5,26]. In this sense, CBL can be viewed as a form of context augmentation that is more explicitly grounded in learning from action.

**2.5 Curriculum Learning and Progressive Capability Development**

Curriculum Learning was first proposed by Bengio and colleagues[35]. Its core idea is to mimic the human educational process by allowing models to learn progressively from easier to more difficult tasks. This strategy has been widely applied in deep learning training, reinforcement learning task scheduling, and multi-stage skill development. In cognitive science, it also corresponds to Vygotsky's concept of the Zone of Proximal Development, which suggests that learners improve most efficiently when faced with appropriately challenging tasks[36].

The curriculum scheduler (M_C) in this work draws on this idea by organizing the injection of historical experience according to case difficulty, thereby helping the agent absorb knowledge more stably. Although the current experimental scale is still limited, this design points to an important direction for future work on progressively cultivating expert-level agents.

**2.6 Positioning and Distinction of This Work**

Overall, this work lies at the intersection of several research directions: LLM-based autonomous agents, long-term memory mechanisms, case-based learning, context-augmented reasoning, and curriculum learning[5,25]. At the same time, it differs from existing approaches in important ways. Compared with conventional agent research, we focus on long-term professional growth rather than single-task execution. Compared with memory architecture research, we emphasize structured experience organization and knowledge transfer[19,20]. Compared with RAG, we reuse practical experience rather than external documents. Compared with few-shot prompting, we distill experience into reusable knowledge rather than simply presenting example answers. Compared with traditional CBR, we study the evolution of professional agents rather than static case matching.

The central contribution of this paper, therefore, is the proposal of a case-based learning mechanism for the formation of real-world professional capability, enabling autonomous agents to learn through practice, grow through cases, and continuously evolve through experience transfer, much like human experts.

**3. Case-Based Learning (CBL) Framework**

**3.1 Design Rationale and Overall Workflow**

The core goal of the Case-Based Learning (CBL) framework proposed in this paper is not simply to provide LLM-based agents with more contextual information, but to build a learning mechanism that more closely resembles the way human experts develop. In real scientific research, engineering practice, and complex system management, professional competence is rarely acquired directly from reading static knowledge. Instead, it is gradually formed through the ongoing process of solving real problems in practice. After completing a task, experts typically do not focus only on the final outcome; they also reflect on which operational paths were effective, which errors should be avoided, which procedures are worth reusing, and which lessons can be transferred to future problems. These summarized practical experiences gradually become professional knowledge.

In contrast, most existing LLM agent frameworks still treat each task execution as an isolated event. After a task is completed, the effective workflows discovered during execution, the methods used for error recovery, and the lessons learned from failure cases are usually discarded rather than transformed into reusable knowledge assets for future tasks. As a result, even when agents repeatedly encounter similar problems, they often need to rediscover solution paths from scratch, lacking any true capacity for cumulative experience.

To address this limitation, CBL enables agents to actively learn by completing real-world cases and to convert what they learn into reusable and transferable professional knowledge for future tasks. In the overall workflow, when an agent receives a new task, the system first retrieves relevant experience from an existing case memory bank, including domain background knowledge, historical failure lessons, verified successful skills, and ordered learning cases. These retrieved experiences are then assembled together with the current task description into an enhanced prompt context, which is provided to the LLM agent for task execution. After the task is completed, the system further analyzes the execution process to identify which strategies succeeded, which errors occurred, and which rules are worth preserving, and then updates the memory modules accordingly to support future tasks. Through this closed loop of case execution → experience summarization → memory consolidation → reuse, each completed real-world case produces a learning increment, allowing the agent to gradually develop more stable professional capability.

## 3.2 Four Types of Structured Experience Memory Modules

To ensure that case experience can be stored reliably and reused effectively, CBL organizes the knowledge acquired by the agent into four complementary structured memory modules, each capturing a different component of professional capability. Fixed Domain Knowledge (M_F) stores stable background knowledge in the task domain, such as API specifications, data structure definitions, platform constraints, and system rules, providing the agent with a basic understanding of how the professional environment operates. System Prompt Constraints (M_S) records behavioral rules distilled from historical failure cases, such as avoiding premature interface calls, terminating invalid loops, or rechecking formats after schema validation failure. By accumulating these constraints, the agent can avoid repeating common errors and execute tasks more robustly.

In addition, Skill Library (M_K) abstracts repeatedly successful operations into reusable method templates, such as configuring vector database workflows, performing schema validation, or generating structured evaluation reports. These skills can then be invoked when similar subtasks appear in future tasks. Finally, Curriculum Organizer (M_C) introduces the idea of curriculum learning by organizing past cases according to difficulty[9]. Easier and more reliable cases are provided first, followed by more complex ones, allowing the agent to absorb historical experience in a more stable and progressive manner. Together, these four modules enable CBL to preserve, organize, and reuse practical experience more effectively than unstructured memory approaches.

**3.3 Experience Update Mechanism**

A key advantage of CBL over static prompting methods is that, after completing each case, the agent genuinely learns something new. To realize this capability, the system introduces an experience update function, $\varphi$ (the memory update function), which automatically summarizes learning outcomes after each task.

Specifically, φ first analyzes execution logs to detect failure patterns that occur during the task, such as timeout, runtime error, blank output, loop detection, and schema validation failure. The system then classifies these errors into standardized failure types and automatically generates new behavioral suggestions based on their causes. For example, when schema validation failure is detected, the system may extract a rule such as "field formats should be rechecked." Finally, these newly acquired experiences are written into the corresponding memory modules, thereby updating the existing knowledge base.

### 3.4 Experience Transfer and Cross-Agent Knowledge Sharing

One of the most important extensions of CBL beyond existing memory architectures is its emphasis on the transferability of experience. Because all learning outcomes are structurally represented as natural-language knowledge modules, the acquired experience is not limited to the original learner, but can be directly read and used by other agents. This means that pitfalls encountered by one agent in real cases can be avoided in advance by others; successful skills mastered by one agent can be directly inherited by others; and behavioral rules summarized by one agent can become shared professional knowledge for the entire group.

In other words, what CBL produces is not merely individual memory, but a form of professional knowledge asset that is shareable and inheritable. This mechanism is particularly important for building expert-level agents in complex domains, because it allows improvements in autonomous agent capability to emerge not only from isolated exploration by individual agents, but also from collective knowledge accumulation through experience transfer.

### 4. Experimental Setup

### 4.1 Task Design and Case Construction

To systematically evaluate the analytical ability and transferability of agents in complex real-world tasks, we construct a case set consisting of six representative task categories. These tasks are drawn from high-complexity scenarios in real system development and deployment, requiring agents not only to provide answers, but also to perform structured diagnosis, causal analysis, and executable solution generation. Unlike traditional single-turn QA benchmarks[16,17], our tasks emphasize whether the analysis process is complete, whether key constraints are respected, and whether the agent can produce repair or governance strategies that are useful in real workflows.

The six task categories cover several major paradigms in current LLM applications, including research-assistant tool orchestration, RLHF training analysis[37-39], enterprise multi-source RAG with access control[5], browser-code-database long-horizon operations, online preference learning with staged review drift, and multi-agent scientific discovery platforms[40,41]. These tasks differ substantially in structure, constraints, and evaluation criteria, forming a heterogeneous benchmark centered on complex scenarios. Some tasks are relatively structured engineering problems with clear evaluation standards, while others are multi-stage, multi-constraint, and more open-ended, allowing us to test the effectiveness of case-based learning across different levels of complexity.

Each task category is instantiated as an independent case built around realistic failure modes. For example, tool-orchestration tasks involve issues such as looping calls, budget overruns, or parameter errors; RLHF tasks include reward improvement with declining human preference, KL drift, or training instability; and multi-agent scientific discovery tasks involve shared memory, novelty constraints, repeated-exploration suppression, and external benchmark validation. In this way, the model is evaluated on realistic analytical requests with clear engineering context rather than on abstract problems. All tasks follow a unified evaluation protocol: each task includes three test rounds, and each round contains five prompt variants, resulting in fifteen evaluation samples per task. This design reduces the influence of prompt phrasing and provides a more stable estimate of task-level performance.

### 4.2 Methods and Baselines

To evaluate the effectiveness of case-based learning, we compare our method with four representative baselines that reflect common LLM analysis paradigms. The first is Zero-Shot, which relies entirely on the model's pretrained capabilities[42]. The second is Few-Shot, which provides a small number of examples to guide structured output generation[4]. The third is Checklist Prompting, which decomposes the task into fixed steps to constrain the reasoning process. The fourth is Rule Memory, which explicitly injects general rules to improve output consistency and stability.

Our method differs from these baselines by introducing case experience, allowing the model to analyze new tasks using knowledge structures accumulated from prior tasks. These experiences are not simple example concatenations, but are organized in structured form, including task-relevant knowledge, analytical prompt templates, and reusable operational skills.

### 4.3 Evaluation Metrics and Success Criteria

Because the six task categories differ in objectives and constraints, we adopt a multi-level evaluation framework consisting of task-specific primary metrics and shared auxiliary metrics. At the task level, each case type is assigned a core metric directly aligned with its objective. For instance, research-assistant tool orchestration is evaluated by task completion and whether budget and schema constraints are satisfied; RLHF training analysis emphasizes held-out preference accuracy, training stability, and the absence of NaN or KL anomalies; enterprise multi-source RAG focuses on permission violations, citation consistency, and grounding quality; and multi-agent scientific discovery is evaluated by real gains on external benchmarks.

At the general level, we additionally report shared statistical metrics across tasks, including mean score, success rate, mean reasoning time, mean token usage, and error count. These metrics capture not only task performance, but also efficiency and system stability. A task is counted as successful only when both its primary metric and key auxiliary constraints meet predefined thresholds. In other words, the model must do more than generate a superficially plausible answer; it must also satisfy the task's core constraints and governance requirements. This makes the evaluation more closely aligned with real-world system standards.

**4.4 Experimental Procedure and Implementation Details**

All methods are evaluated under a unified procedure. For each test sample, the model receives the task description and input prompt, then produces a complete analysis and solution. After normalization, the output is passed to the evaluation module, which computes both task-specific and shared metrics.

To ensure fair comparison, the experiments tightly control several factors. All methods use the same model configuration and inference parameters. The prompt protocol is kept consistent across methods to preserve sample-level comparability. The evaluation scripts are fully shared across all settings, ensuring a unified scoring standard. In addition, reasoning time and token consumption are recorded to analyze computational cost differences between methods. Overall, this experimental framework balances task complexity, methodological comparability, and evaluation consistency, providing a reliable basis for the results analysis. Importantly, the six task categories are not simply assembled together; they form a progressive evaluation spectrum ranging from structured engineering tasks to highly complex open-ended tasks.

**5. Results**

## 5.1 Overall Results Across the Six Tasks

Figure 1 illustrates the proposed Case-Based Learning (CBL) framework and its core operating mechanisms. Unlike approaches that enhance LLM agents merely by increasing context length or injecting static knowledge, the central idea of CBL is to treat each real task execution as a learnable case. In this way, the agent can gradually accumulate experience through the continual resolution of real problems and develop along a trajectory more similar to that of human experts. Specifically, when the system receives a new task, it first retrieves relevant historical experience from an existing case memory and combines this experience with the current task description to form an enhanced prompt context for the LLM agent to perform analysis and decision-making. After the task is completed, the system further summarizes effective strategies, lessons from failure, and reusable procedures from the current case, and stores them as knowledge that can be invoked again in future tasks. This creates a continuous closed loop of case execution → experience summarization → memory consolidation → reuse.

To support stable storage and effective reuse of case experience, CBL organizes learned knowledge into four complementary structured memory modules. First, the Fixed Domain Knowledge module ($M\_F$) stores stable background knowledge in the task domain, such as interface specifications, data structure definitions, and system constraints, thereby providing the agent with foundational domain understanding. Second, the System Prompt Constraints module ($M\_S$) records behavioral rules distilled from historical failure cases, helping the agent avoid repeating previous mistakes. Third, the Skill Library module ($M\_K$) stores method templates that have been repeatedly validated across multiple cases, allowing successful strategies to be directly reused. Finally, the Curriculum Organizer module ($M\_C$) orders historical experience according to case difficulty and success rate, enabling the system to first absorb easier and more stable experience before gradually extending to more complex tasks.

Another key feature of CBL is its explicit experience update mechanism. After each task, the system analyzes execution logs and trajectories, automatically extracts success patterns and failure patterns, and further abstracts them into reusable behavioral rules, skill templates, or curriculum experience, which are then written back to the corresponding memory modules. Thus, when an agent completes a case, it is not merely finishing a task; it is transforming the execution process itself into knowledge resources for future problem solving. Through this mechanism, CBL moves from one-off task completion to continual learning.

More importantly, the experience consolidated by CBL is not confined to the original learner, but can also be transferred across agents. Because the system represents learned results as structured natural-language knowledge modules, the rules, skills, and case experience summarized by the original agent from real tasks can be directly inherited and reused by other agents. This means that knowledge formed by one agent in practice can become a shared professional asset for the entire group, allowing capability improvement in multi-agent systems to emerge not from repeated exploration by individual agents alone, but from more efficient co-evolution through experience transfer.

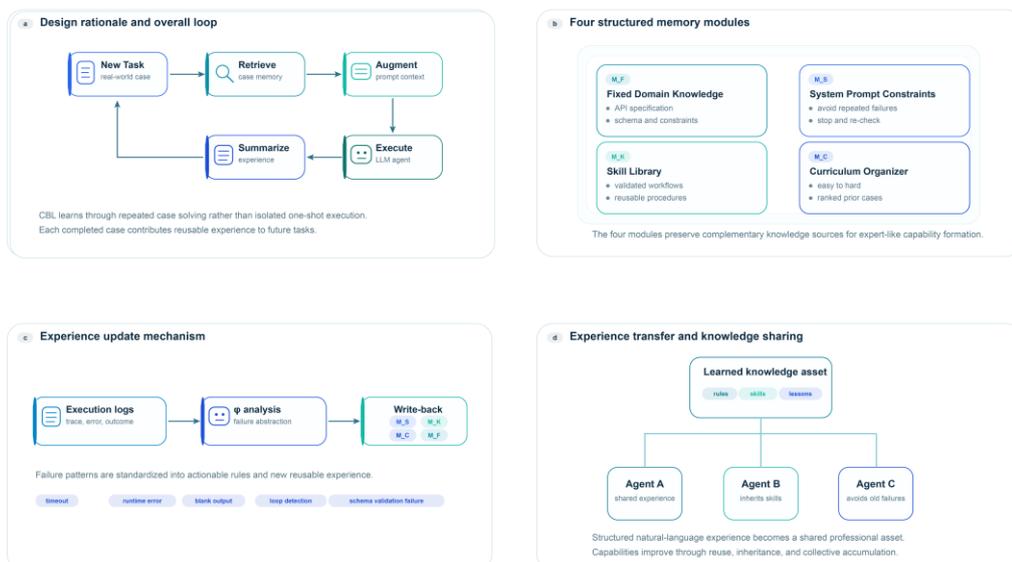

**Figure 1 | The Case-Based Learning (CBL) framework: a continual case-based learning and experience transfer mechanism for LLM agents.** (A) Design rationale and workflow. CBL treats each real task as a learnable case. For a new task, the system retrieves relevant prior experience, combines it with the task description, and uses the resulting prompt to guide agent execution. Afterward, successful strategies, failure lessons, and reusable procedures are summarized and stored, forming a continual loop of execution, summarization, consolidation, and reuse. (B) Four structured memory modules. CBL organizes knowledge into four complementary components: Fixed Domain Knowledge (M_F), System Prompt Constraints (M_S), Skill Library (M_K), and Curriculum Organizer (M_C), corresponding to domain knowledge, error avoidance, reusable skills, and learning order. (C) Experience update mechanism. Using the explicit update function φ, the system analyzes execution logs, extracts success and failure patterns, abstracts them into reusable knowledge, and writes them back to the appropriate memory modules. (D) Experience transfer and cross-agent sharing. Because learned experience is represented as structured natural-language knowledge, the rules, skills, and case experience acquired by one agent can be directly inherited and reused by others, turning individual experience into shared professional assets.

## 5.2 Overall Results Across the Six Tasks

We first perform an overall evaluation of the proposed case-based learning method across six task categories: research-assistant tool orchestration, RLHF training analysis, enterprise-grade multi-source RAG with access control, long-horizon operations, online preference learning with staged review drift, and multi-agent scientific discovery platforms. Together, these tasks form a continuum from structured engineering problems to highly complex open-ended tasks. Overall, our method achieves score_mean = 4.0 and success_rate = 1.0 on all six tasks, with an average score of 4.0, average success rate of 1.0, mean reasoning time of approximately 25,507.6 ms, and mean token usage of approximately 129,049.9. These results indicate that case-based learning not only maintains highly consistent and stable performance across diverse tasks, but also enables the model to reach perfect-level task completion quality on this six-task benchmark.

Compared with baseline methods, our approach matches or exceeds the best baseline on all six tasks. On research-assistant tool orchestration and RLHF training analysis, our method outperforms the best baseline in score while matching it in success rate. On enterprise-grade multi-source RAG with access control, long-horizon operations, online preference learning with staged review drift, and multi-agent scientific discovery, our method outperforms the best baseline in both score and success rate. Overall, case-based learning not only preserves stable high performance on structured tasks, but further expands its lead over baselines as task complexity increases (Figure 2).

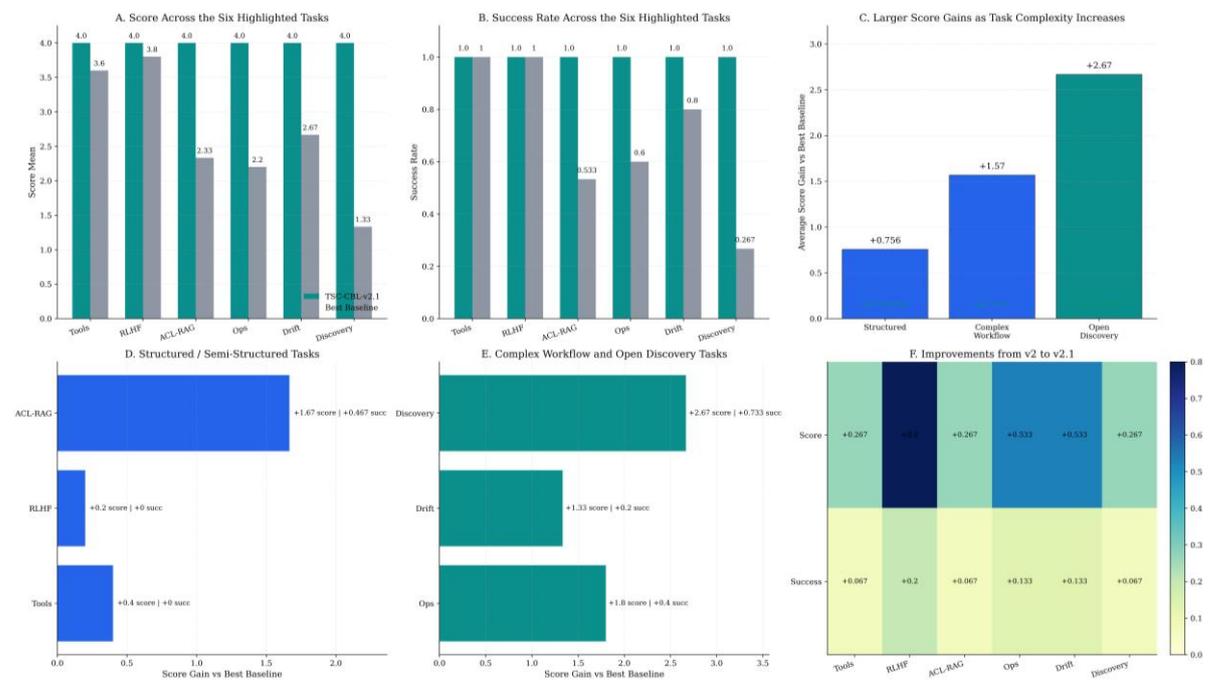

**Figure 2. Overview of overall six-task results and complexity-dependent gains.** This figure summarizes the overall performance of the proposed method across the six task categories. Panels A and B compare our method with the best baseline in terms of score_mean and success_rate, respectively. Panel C shows how the average gain over the best baseline changes as tasks progress from structured tasks to complex workflow tasks and then to open-ended discovery tasks. Panels D and E further break down the margin of improvement in structured/semi-structured tasks and in complex workflow/open-ended discovery tasks. Panel F shows the overall gain of the present method relative to the previous version.

More importantly, these results do not merely show that "case-based learning works." Rather, they reveal a stronger pattern: the task-internal experience accumulated through case-based learning continually strengthens the model's structured analytical capability, and this benefit becomes larger on more complex tasks.

## 5.3 Results on Structured and Semi-Structured Tasks

Among the six tasks, research-assistant tool orchestration, RLHF training analysis, and enterprise-grade multi-source RAG with access control can be regarded as relatively well-structured tasks. They share strong engineering-process characteristics, clear evaluation targets, and key scoring signals that are more easily covered through structured analysis. This group of tasks is therefore used primarily to test whether case-based learning can stably improve the model's ability to capture task structure and key constraints.

On the research-assistant tool orchestration task, our method achieves score_mean = 4.0 and success_rate = 1.0. The best baseline attains a score of 3.6 and a success rate of 1.0, meaning that our method improves the score by +0.4 while maintaining tied-best success rate.

This suggests that once termination conditions, budget control, schema validation, and loop recovery are organized into an explicit task-specific analytical scaffold, the model can more reliably produce complete solutions that cover the key governance points. On the RLHF training analysis task, our method again achieves 4.0 / 1.0, whereas the best baseline, Checklist Prompt, achieves 3.8 / 1.0. Thus, our method maintains a score advantage of +0.2 while tying for the best success rate. This result is particularly important because RLHF training analysis requires the model to handle multiple high-value signals simultaneously, including held-out/OOD behavior, KL/NaN anomalies, reward hacking, and rollback. Case-based learning helps ensure that these critical diagnostic points are not overlooked.

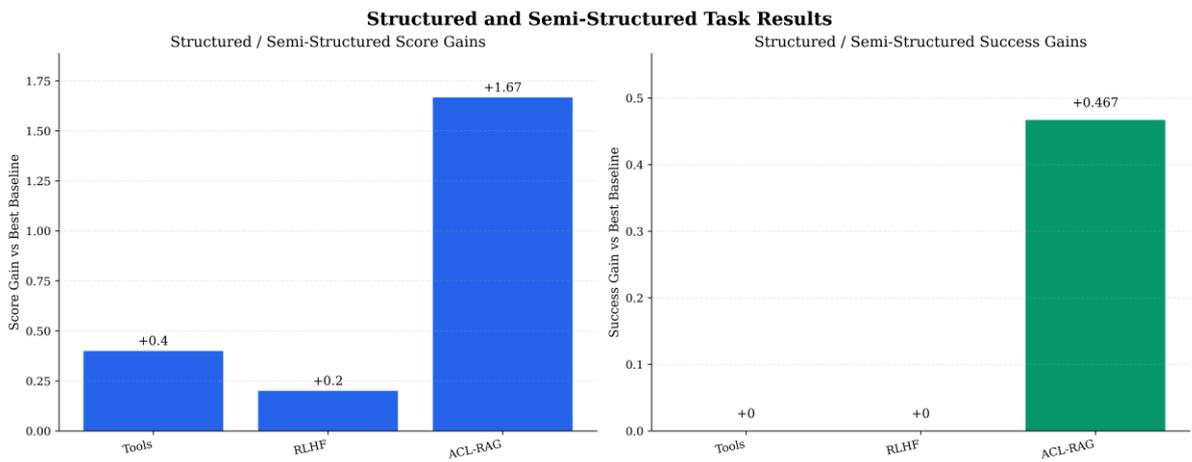

**Figure 3. Performance gains on structured and semi-structured tasks.** This figure shows the improvements achieved by the proposed method on three structured or semi-structured tasks: research-assistant tool orchestration, RLHF training analysis, and enterprise-grade multi-source RAG with access control. The left panel shows the score improvement over the best baseline, and the right panel shows the improvement in success rate.

On the enterprise-grade multi-source RAG with access control task, our method achieves 4.0 / 1.0, whereas the best baseline reaches only 2.333 / 0.533. In other words, our method obtains a score advantage of +1.667 and a success-rate advantage of +0.467 on this task. This indicates that case-based learning is particularly effective for multi-constraint problems involving proactive security boundary enforcement, evidence traceability consistency, and audit-chain completeness, because it helps the model construct a unified analytical structure spanning retrieval, caching, citation handling, and final answer generation (Figure 3). Taken together, these three tasks show that case-based learning does not merely allow the model to "know more rules"; rather, it enables the model to organize task analysis more reliably and to cover the structured signals emphasized by the evaluator more completely.

**5.4 Stronger Gains on Complex Workflow Tasks**

If the previous three tasks mainly demonstrate that case-based learning can stably improve coverage quality on structured engineering problems, the long-horizon operations task and the online preference learning with staged review drift task further show that the gains become even more pronounced as task complexity increases. The long-horizon operations task requires the model to handle multi-environment switching, state refreshing, checkpoint setting, rollback gating, and replanning after side-effect actions. This is not a simple matter of "listing steps"; it requires the model to continuously maintain correct judgment of system state and risk boundaries throughout a multi-stage operation. On this task, our method achieves score_mean = 4.0 and success_rate = 1.0, while the best baseline reaches only 2.2 / 0.6. Thus, our method improves score by +1.8 and success rate by +0.4. This result shows that, in long-chain, high-side-effect, and highly constrained tasks, case-based learning can substantially strengthen the model's ability to preserve key control conditions.

The online preference learning with staged review drift task shows a similar trend. This task requires the model not only to handle judge drift, calibration evaluation, and hierarchical gating, but also to move beyond overall average metrics and explicitly attend to risks affecting key groups and worst-case groups. On this task, our method achieves 4.0 / 1.0, whereas the best baseline achieves 2.667 / 0.8. Accordingly, our method improves score by +1.333 and success rate by +0.2 (Figure 4). This indicates that when the task requires multi-gate governance rather than single-metric optimization, case-based learning provides a more effective structural prior, allowing the model to remain highly stable under complex decision criteria.

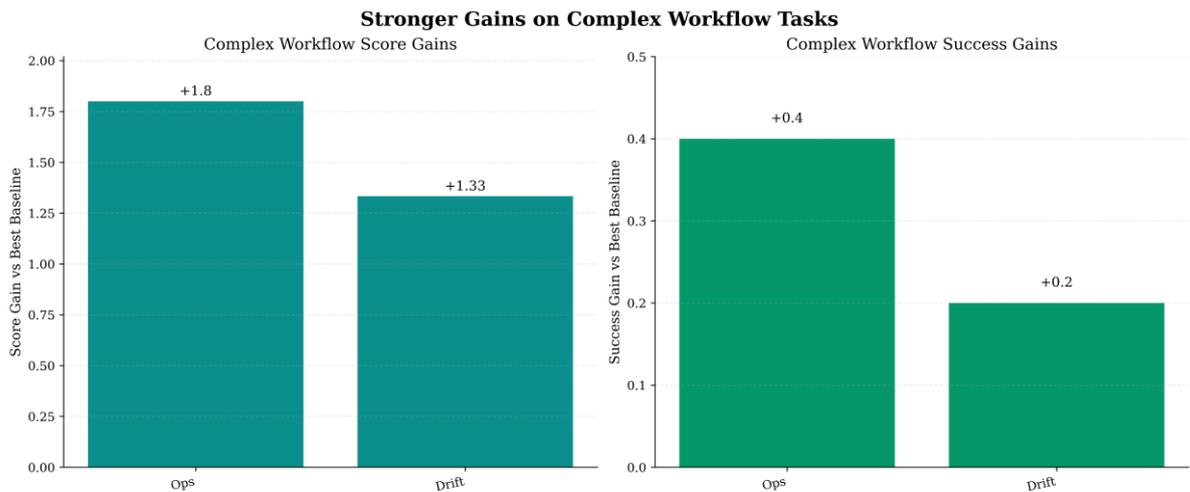

**Figure 4. Stronger gains on complex workflow tasks.** This figure shows the improvements of the proposed method on two complex workflow tasks: long-horizon operations and online preference learning with staged review governance. The left panel shows the score gain over the best baseline, and the right panel shows the gain in success rate. As shown, when tasks require the model to continuously handle multi-stage constraints such as state refreshing, checkpoint setting, rollback gating, hierarchical evaluation, and worst-group governance, the advantage of case-based learning becomes more pronounced.

Together, these two tasks reveal an important pattern: as tasks shift from static and clearly defined analysis problems to dynamic, multi-stage, and strongly constrained workflows, the task-internal experience accumulated by case-based learning becomes not merely beneficial, but increasingly decisive for whether the model can succeed reliably. In other words, the more complex the task, the greater the value provided by case-based learning.

**5.5 Largest Advantage on the Most Open-Ended Discovery Task**

Among the six tasks, the multi-agent scientific discovery platform is the most open-ended and complex, and thus most clearly reflects the upper bound of the case-based learning approach. Compared with the preceding tasks, this task requires the model not only to produce an analysis plan, but also to simultaneously handle shared memory, novelty constraints, budget allocation, external benchmark validation, and suppression of repeated exploration. What it evaluates is not merely local diagnostic ability, but whether the model can maintain an overall discovery-governance structure in an open problem space.

On this task, our method achieves score_mean = 4.0 and success_rate = 1.0. In contrast, the best baseline attains only 1.333 in score and 0.267 in success rate. Therefore, our method yields a score advantage of +2.667 and a success-rate advantage of +0.733, the largest improvement observed across all six tasks (Figure 5). This result has clear methodological significance. The multi-agent scientific discovery task cannot be solved reliably through simple templates or shallow rules, because it requires the model to manage shared memory across multiple sub-agents, control novelty, constrain repeated exploration, and determine whether discoveries are genuinely valid using external benchmarks rather than internal bootstrap metrics. The strong performance of our method on this task shows that case-based learning provides not just ordinary prompt enhancement, but a structured experience-transfer mechanism capable of supporting open-ended and complex discovery processes.

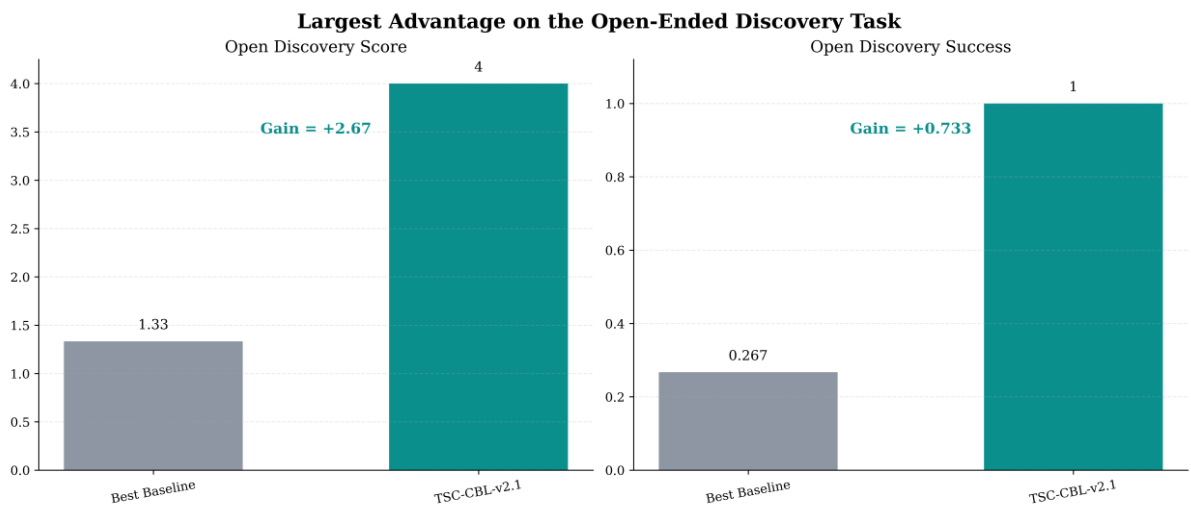

**Figure 5. Largest advantage on the open-ended scientific discovery task.** This figure focuses on the multi-agent scientific discovery platform task and compares the proposed method with the best baseline in terms of score and success rate. The results show that our method achieves the most substantial margin of improvement on this task, indicating that case-based learning not only enhances general analytical ability, but also provides effective structured experiential support for shared-memory management, novelty constraints, budget allocation, external benchmark validation, and repeated-exploration suppression in open problem spaces.

## 5.6 Transferability of Practical Knowledge

A further question is whether the benefit of case-based learning is limited to the original agent itself, or whether the practical knowledge formed by the agent in real tasks can be externalized and transferred to other agents. If the latter is true, then the significance of case-based learning lies not only in improving a single agent's capability, but also in providing inheritable professional experience for subsequent agents, thereby establishing a cross-agent mechanism of knowledge accumulation.

To answer this question, we conduct a dedicated analysis of the practical knowledge accumulated across the six projects. Rather than directly reusing the original execution trajectories, we reorganize the experience from the six projects into four reusable modules, as shown in Figure 6A. Specifically, the case-based learning framework represents knowledge as Fixed Domain Knowledge (M_F), System Prompt Constraints (M_S), Skill Library (M_K), and Curriculum Organizer (M_C). Here, M_F stores stable domain background and system constraints, M_S records behavioral rules abstracted from failure cases, M_K stores method templates validated across multiple cases, and M_C organizes the order in which experience is invoked according to case difficulty and success rate. In this representation, what is inherited by a new agent is not the superficial form of original dialogues or trajectories, but the structured practical knowledge distilled from real project execution.

After inheriting these four modules, the new agent exhibits consistently high and stable performance across all six target tasks. As shown in Figure 6B, on research-assistant tool orchestration, RLHF training analysis, enterprise-grade multi-source RAG with access control, long-horizon operations, online preference learning with staged review drift, and multi-agent scientific discovery platforms, the inherited new agent achieves score_mean = 4.0 and success_rate = 1.0 on all six tasks. This result shows that the experience formed by the original agent in real projects can not only be explicitly preserved, but can also be directly reused by a new agent after reorganization, supporting uniformly high performance across multiple same-domain but structurally different tasks.

Module ablation results provide more direct evidence of the underlying mechanism. As shown in Figure 6C, on two transfer-diagnostic tasks, removing the Fixed Domain Knowledge module (M_F) reduces the average score to 3.0 and the success rate to 0.9. Removing the System Prompt Constraints module (M_S) lowers the average score to 3.3. Removing the Skill Library module (M_K) lowers the average score to 3.4. Removing the Curriculum Organizer module (M_C) lowers the average score to 3.2. By comparison, the non-case-based Rule Memory baseline achieves only an average score of 2.9. These findings indicate that the inherited capability is not driven by a single prompt, a single rule, or accidental template matching; rather, it depends on the coordinated support of multiple types of practical knowledge modules. If any one of the four modules is missing, transfer performance declines substantially.

Analysis of key behavioral signals further shows that this transfer is not merely a generic strengthening, but an inheritance of structured behavioral experience. As shown in Figure 6D, in the cross-lingual RAG task, removing M_F reduces the trigger rate of the "avoid_premature_tuning" signal to 0.2, indicating that fixed domain knowledge directly affects whether the agent prioritizes the correct domain-diagnostic order. In the research-assistant agent task, removing M_F also causes the "termination" signal to drop from full coverage to 0.8. Meanwhile, removing M_S, M_K, and M_C respectively affects different types of signals, including termination rules, method templates, and the order of experience invocation. This shows that the four modules preserve different dimensions of practical knowledge, and together they determine whether a new agent can reproduce the high-value behavioral patterns supported by prior experience。

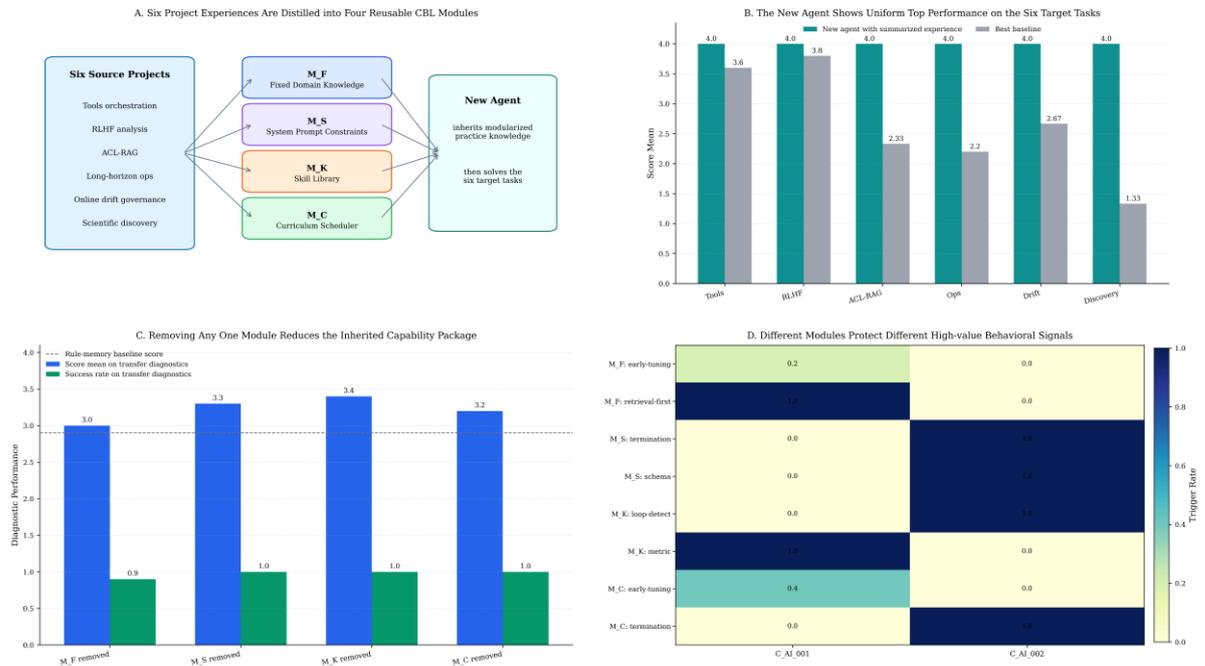

**Figure 6. Analysis of the inheritability of practical knowledge.** (A) Experience from the six projects is not directly reused in raw trajectory form, but is reorganized into four complementary modules: Fixed Domain Knowledge (M_F), System Prompt Constraints (M_S), Skill Library (M_K), and Curriculum Organizer (M_C), corresponding respectively to domain cognition, error avoidance, successful skills, and learning order. (B) After inheriting these modules, the new agent achieves uniformly high performance across all six target tasks, showing that modularized practical knowledge can be stably reused across tasks. (C) Module ablation results show that removing any of the four modules—fixed domain knowledge, system prompt constraints, skill library, or curriculum organizer—reduces inherited capability, indicating that the new agent's performance depends on the coordinated effect of multiple practical knowledge modules. (D) Trigger rates of key behavioral signals further show that the four modules protect different types of high-value behavioral patterns, rather than redundantly encoding the same capability.

Taken together, these results support the conclusion that the practical knowledge formed by one agent in real projects can be explicitly organized into four complementary modules—M_F, M_S, M_K, and M_C—and can continue to function after being transferred to a new agent. What the new agent inherits is not the original execution process itself, but the abstracted domain understanding, failure-avoidance rules, successful skill templates, and ordering of experience invocation. Therefore, the key contribution of case-based learning lies not only in improving single-task outcomes, but also in transforming real-world practical experience into inheritable, reusable, and composable capability assets.

## 6. Discussion

The experimental results show that the core value of case-based learning is not simply to provide LLMs with more background information, but to equip agents with a learning mechanism that more closely resembles the growth process of real experts. Unlike approaches that rely on pretrained knowledge, prompt engineering, or few-shot examples[4,5], case-based learning emphasizes having agents work on real tasks, analyze real problems, and accumulate experience through real failures and real repairs. As a result, the object of learning is no longer limited to abstract rules in static text, but extends to practical knowledge derived from real-world workflows. This is the fundamental difference between case-based learning and conventional knowledge-augmentation methods: the former focuses on the formation and transfer of experience, whereas the latter relies mainly on recombining and invoking existing textual knowledge.

Across the six task categories, the benefits of this learning mechanism are not uniform, but become more pronounced as task complexity increases. On structured or semi-structured tasks such as research-assistant tool orchestration, RLHF training analysis, and enterprise-grade multi-source RAG with access control, case-based learning already achieves consistently optimal or tied-optimal performance, indicating that it effectively improves the model's coverage of key constraints and high-value scoring signals. On more complex tasks, such as long-horizon operations, online preference learning with staged review drift, and multi-agent scientific discovery platforms, the margin of improvement becomes even larger. This suggests that the more complex, open-ended, and multi-stage a task is, the more task-internal experience accumulated through case-based learning can be converted into meaningful performance gains. This trend is also consistent with the broader intuition behind curriculum-based capability development, namely that increasingly demanding tasks make differences in internal knowledge organization more visible[35].

This pattern has clear methodological implications. In relatively well-structured tasks, models can often cover most key points using checklists, rule memory, or explicit procedural templates, so the advantage of case-based learning is mainly reflected in more stable high scores and more complete structured outputs. In complex tasks, however, the real challenge is usually not whether the model "knows a rule," but whether it can invoke the right experience at the right time and organize its reasoning around the most critical constraints. For example, in long-horizon operations, the model must refresh state after side-effect actions and immediately stop or roll back when a key invariant fails. In multi-agent scientific discovery, it must simultaneously manage shared memory, novelty constraints, budget allocation, and external validation. For such problems, static textual knowledge alone is rarely sufficient to support complete and stable reasoning, whereas the task experience accumulated through case-based learning can provide a higher-level structural prior. More broadly, this interpretation is aligned with prior work suggesting that long-context or retrieval-based augmentation alone does not guarantee effective use of relevant information during reasoning[3,28].

The results also suggest that case-based learning can form a high-value collaborative relationship with human domain experts. Case-based learning does not imply replacing experts entirely; rather, it offers a mechanism for amplifying expert experience. If genuine domain experts participate in the case-learning process, they do not need to repeatedly provide complete answers for every task. Instead, they can contribute high-value guidance on key cases, for example by identifying the real root cause, correcting an analytical path, adding critical constraints, or confirming which experiences are worth preserving and transferring. The agent can then consolidate these judgments into reusable case experience and continue to invoke them in future tasks. In this way, the expert's role shifts from repeatedly solving individual problems to providing high-value guidance and knowledge calibration on pivotal cases.

Overall, the present work supports the view that case-based learning is not merely a supplement to static text-based knowledge learning, but a way for agents to work, learn, and evolve through real-world tasks. Its value lies not only in the stable gains observed across the current six tasks, but also in providing a developmental path for future professional agents that is more aligned with the way real experts actually grow.

## 7. Limitations and Future Work

Although our results show that case-driven experience transfer has clear potential for improving both performance on complex tasks and reasoning efficiency, the current work still has several limitations. First, the organization of case assets remains relatively static. In this study, experience is represented as structured assets such as fixed knowledge, system prompts, and operational skills. While this improves interpretability, the overall experience structure still depends largely on predefined categories and has not yet achieved more dynamic and fine-grained automatic updating. Second, the modeling of relationships between tasks remains insufficient. At present, experience transfer occurs mainly at relatively shallow levels of knowledge and skills; a stronger task graph or hierarchical experience system has not yet been established. As a result, transfer may be limited when tasks are only partially similar but differ substantially in constraint structure. Third, although the six task categories cover several representative scenarios, the overall benchmark scale is still limited and is better viewed as an initial validation of methodological effectiveness rather than a universal conclusion for all agent scenarios. Finally, the current statistical analysis focuses mainly on trend comparison and effect observation, and does not yet incorporate a more rigorous large-scale significance testing framework.

Future work can proceed in four directions. First, more automated mechanisms for experience extraction and updating can be developed so that the case library continuously expands during task execution. Second, finer-grained strategies for experience routing and task matching can be explored to improve transfer precision[24,25]. Third, case-based learning can be integrated with explicit state modeling, multi-agent collaboration, and external tool supervision to enhance practical execution ability in complex dynamic environments. Fourth, the benchmark can be expanded further and paired with stricter statistical validation to provide a stronger empirical foundation for case-based learning methods[40,41].

## 8. Conclusion

The results of this study show that the value of case-based learning lies not merely in providing LLMs with more prompts or background knowledge, but in giving agents a learning mechanism that more closely resembles the growth process of real experts. Unlike methods that rely on pretrained knowledge, prompt engineering, or few-shot examples, case-based learning emphasizes having agents work on real tasks, analyze real problems, and accumulate experience through real failures and repairs. As a result, the object of learning is no longer limited to abstract rules in static text, but extends to practical knowledge drawn from real-world workflows.

Across the six task categories, this learning mechanism is especially effective on complex tasks. On relatively well-structured tasks, case-based learning already achieves optimal or tied-optimal performance; on more complex tasks, such as long-horizon operations, staged-review governance, and multi-agent scientific discovery, its advantages become even larger. This suggests that the more complex, open-ended, and multi-stage a task is, the more the experience formed through case-based learning can be translated into performance gains. In other words, the key challenge in complex tasks is not simply "knowing the rules," but whether the model can invoke the right experience at the right time and organize reasoning around the critical constraints. Overall, case-based learning is not merely a supplement to static text-based learning, but a way for agents to work, learn, and evolve in real-world tasks. This mechanism not only improves performance on complex tasks, but also offers a development path for future professional agents that is more consistent with how real experts actually grow.

**Appendix**

**Appendix A. Additional Details of the Eight Benchmark Tasks**

The six task categories in this study are not ordinary question-answering samples, but a collection of cases built around high-complexity problems commonly encountered in real system development, analysis, and governance. Their shared characteristic is that they do not have a single standard answer; instead, they require the agent to produce structured and executable analyses around symptoms, root causes, constraints, and repair paths. To avoid excessive length in the main text, this appendix provides additional details on the background, problem boundaries, and evaluation focus of each task.

A.1 Research-Assistant Agent Tool Orchestration

This task simulates a research-assistant system capable of invoking search, web scraping, PDF reading, and summarization tools. The model must identify issues such as looping tool calls, budget overruns, and parameter format errors in multi-tool workflows, and propose repair strategies with explicit termination conditions and state-management ability. The key question is whether the agent recognizes that tool use is not simply free-form generation, but must follow execution logic that is verifiable, terminable, and controllable.

A.2 Preference Optimization and RLHF Training

This task places the agent in the role of an alignment-training analyst and asks it to explain problems such as rising reward but declining real quality, KL drift in PPO training, and numerical instability. The goal is not to restate the RLHF training pipeline, but to distinguish between superficial training signals and genuine preference improvement, and to propose repair strategies aimed at training stability.

A.3 Enterprise-Grade Multi-Source RAG with Access Control

This task requires the agent to analyze an enterprise knowledge system connected simultaneously to documents, spreadsheets, code repositories, and knowledge bases, while handling issues such as permission leakage, evidence traceability, and incomplete auditing. Unlike ordinary RAG systems, answer quality is not the only objective here; security and evidence consistency are also necessary conditions for success. The model must clearly identify where security boundaries should be established and how citation chains can remain stable.

A.4 Browser–Code–Database Long-Horizon Operations

This task simulates long-chain operations across browser interfaces, code repositories, databases, and deployment systems. Because actions such as clicking, committing, migrating, or deploying have real side effects, the task shifts attention to state refresh, checkpoint management, transaction consistency, and failure rollback. The model must recognize that, in environments with side effects, previous plans may become invalid immediately after a single action and therefore the analytical path must be updated dynamically.

A.5 Online Preference Learning and Staged Review Drift

This task focuses on judge-model drift and staged-release risk in online preference learning systems. The agent must explain why improvement in average metrics does not necessarily mean the system is safe to deploy, and propose analyses centered on hierarchical evaluation, key-group gating, and worst-group risk control. The emphasis of this task is on constraints and governance rather than simple metric interpretation.

A.6 Multi-Agent Scientific Discovery Platform

This task simulates a setting in which multiple sub-agents generate hypotheses in parallel, allocate budget, and conduct scientific exploration. The challenge is not merely to produce more ideas, but to share memory, suppress repeated exploration, and determine whether the system achieves genuine scientific gains using external benchmarks rather than internal bootstrap metrics. This is the most open-ended and complex of the six tasks.

**Appendix B. Baseline Definitions**

To ensure clear comparison semantics, the four baseline methods used in this paper are defined as follows.

Zero-Shot. Only the current task description is provided, without any additional examples, rules, or process prompts. This baseline is used to measure the raw analytical ability of the base model without external experience support.

Few-Shot. A small number of examples are added to the task instruction to simulate in-context learning. This method can provide local guidance, but its knowledge transfer depends mainly on superficial similarity between examples and usually lacks explicit constraints and stable task-structure control.

Checklist Prompt. A manually designed step-by-step checklist is used to guide the model through a predefined analysis process. This method emphasizes structure and process control and is suitable for problems with clear engineering order, but its generality is usually limited by the coverage of the checklist itself.

Rule Memory. General experience constraints are injected into the model in rule form to improve output consistency and controllability. Compared with Checklist Prompt, it is closer to a static experience-memory mechanism, but it usually lacks task-level contextual adaptation ability.